\pdfoutput=1

\documentclass[11pt]{article}

\usepackage{ACL2023}

\usepackage{times}
\usepackage{latexsym}

\usepackage[T1]{fontenc}

\usepackage[utf8]{inputenc}

\usepackage{microtype}

\usepackage{inconsolata}
\usepackage{comment}
\usepackage{graphicx}
\graphicspath{{images/}}
\usepackage{caption}
\usepackage{subcaption}
\usepackage{hyperref}
\usepackage{xcolor}
\usepackage{booktabs}
\usepackage{color, colortbl}

\title{Linguistic representations for fewer-shot relation extraction across domains}

\author{Sireesh Gururaja \and Ritam Dutt \and Tinglong Liao \and Carolyn Rosé \\
  Language Technologies Institute \\ Carnegie Mellon University \\
  \texttt{\{sgururaj, rdutt, tinglonl, cprose\}@cs.cmu.edu}
}

\begin{document}
\maketitle
\begin{abstract}

Recent work has demonstrated the positive impact of incorporating linguistic representations as additional context and scaffolding on the in-domain performance of several NLP tasks. We extend this work by exploring the impact of linguistic representations on cross-domain performance in a few-shot transfer setting. An important question is whether linguistic representations enhance generalizability by providing features that function as cross-domain pivots. We focus on the task of relation extraction on three datasets of procedural text in two domains, cooking and materials science. Our approach augments a popular transformer-based architecture by alternately incorporating syntactic and semantic graphs constructed by freely available off-the-shelf tools. We examine their utility for enhancing generalization, and investigate whether earlier findings, e.g. that semantic representations can be more helpful than syntactic ones, extend to relation extraction in multiple domains. We find that while the inclusion of these graphs results in significantly higher performance in few-shot transfer, both types of graph exhibit roughly equivalent utility.


\end{abstract}

\section{Introduction}

In many specialized domains, such as healthcare or finance, one of the principal limitations for the implementation of machine learning based NLP methods is the availability of high quality data, which tends to be both time-consuming and expensive to acquire. While pre-trained language models allow impressive performance gains across a number of tasks, those gains frequently fail to generalize to specialized domains. Robust representations that allow models to both take advantage of pretrained models and generalize across domains are therefore highly desirable. 

Recent works such as \citet{prange-etal-2022-linguistic} have demonstrated the significant potential of using human-annotated linguistic information as scaffolding for learning language models. Other works such as \citet{zhang-ji-2021-abstract} and \citet{bai-etal-2021-semantic} use automatically generated semantic annotations. These works depend on the idea that the structure that the linguistic frameworks provide allows models to better learn salient features of the input. In addition, however, linguistic structures offer abstraction over the variation of natural language, providing representations that might express meaning in domain-general ways. We therefore extend earlier work to investigate whether including linguistic representations encourages learning domain-agnostic representations of relations such that models can generalize better in a few-shot setting, i.e. learning from less high-quality data in new domains. We focus on the case of automatically generated linguistic annotations, to evaluate the impact they can have on downstream tasks without expensive human annotation of parse data. 

We use two linguistic formalisms, to evaluate and compare their impact: dependency parses, and abstract meaning representations (AMR). AMR \citep{amr} seeks to represent meaning at the level of a sentence in the form of a rooted, directed graph. AMR is based on Propbank \citep{propbank}, and factors out syntactic transformations due to verb alternations, passivization, and relativization, leading to a less sparse expression of textual variance. Dependency parsing, by contrast, remains at a low level of abstraction, with structures that do not nest outside of the words in the original text.

We study the problem of relation extraction on procedural text. We use procedural text because of what we term their \textit{implicit schemas}. Across domains, our datasets describe the process of combining ingredients under certain conditions in a sequential fashion to produce a desired product. These range from preparing a cooking recipe to synthesizing a chemical compound to extracting materials from ores. As a result, the relations that we derive from each dataset share a loose semantic correspondence, both to each other and to basic semantic relations such as verb arguments and locations. 
For example, the actions ``boil'' and ``heat'' in ``Boil the mixture in a medium saucepan'' and ``Heat the solvent in the crucible'' are similar.

We hypothesize that the underlying semantics of all of these datasets are similar enough that models should be able to better generalize across domains from the explicit inclusion of syntactic and semantic structural features. We use three datasets across two domains: recipes for cooking, and materials science synthesis procedures. Each of these datasets defines the task of generating a comprehensive, descriptive graph representation of a procedure. We simplify this task into relation extraction in order to better compare the impact of different linguistic formalisms.

We augment a popular transformer-based relation extraction baseline with features derived from AMR \citep{amr} and dependency parses and investigate their impact in a few-shot setting both in-domain and across domains. Experiments show that both AMR parses and dependencies significantly enhance model performance in few-shot settings but that the benefit disappears when models are trained on more data. We additionally find that while cross-domain transfer can degrade the performance of purely text-based models, models that incorporate linguistic graphs provide gains that are robust to those effects. 

We make our code available with our submission\footnote{\url{https://github.com/ShoRit/flow_graphs}}. 

\section{Related Work}

\subsection{Few-shot Relation Extraction}

The goal of relation extraction (RE) is to detect and classify the relation between specified entities in a text according to some predefined schema. Current research in RE has mostly been carried out in a few-shot or a zero-shot setting to address the dearth of training data  \cite{liu-etal-2022-simple} and the ``long-tail'' problem of skewness in relation classes. \cite{MLMAN_RE}. Salient work in that direction includes 
(i) designing RE-specific pretraining objectives for learning better representations \cite{mtb, zhenzhen-etal-2022-improving,wang-etal-2022-deepstruct}, (ii) incorporating  meta-information such as relation descriptions \cite{TD_Proto, zsbert} a global relation graph, \cite{REGRAB}, or entity types \cite{peng-cp}, and (iii) leveraging  additional information in the form of dependency parses \cite{DAPL}, translated texts for multilingual RE \cite{indore}, or distantly supervised instances \cite{DS-rel-1, DS-rel-2}.  All of these techniques seek to alleviate the need of using expensive human-annotated training data. In this work, we question whether incorporating linguistic structure on existing models can aid learning robust representations which can be transferred to other domains. 

\subsection{Linguistic frameworks for NLP}

Supplementing training data with explicit linguistic structure, in the form of syntactic and semantic parses has led to substantial improvements in the in-domain performance on several NLP tasks. \citet{sachan-etal-2021-syntax} challenges the utility of syntax trees over pre-trained transformers for IE and observed that one can only obtain meaningful gains with gold parses. Semantic parses, in the form of AMRs, have shown to be beneficial for IE \cite{bioamr, zhang-ji-2021-abstract, amr-docs}, even when the parses employed are not human-annotated. In this work, we raise the question of the utility of either kind of parse for few-shot RE in a cross-domain setting.

\section{Methodology}

\begin{figure}[ht]
\centering
    \includegraphics[width=0.8 \linewidth]{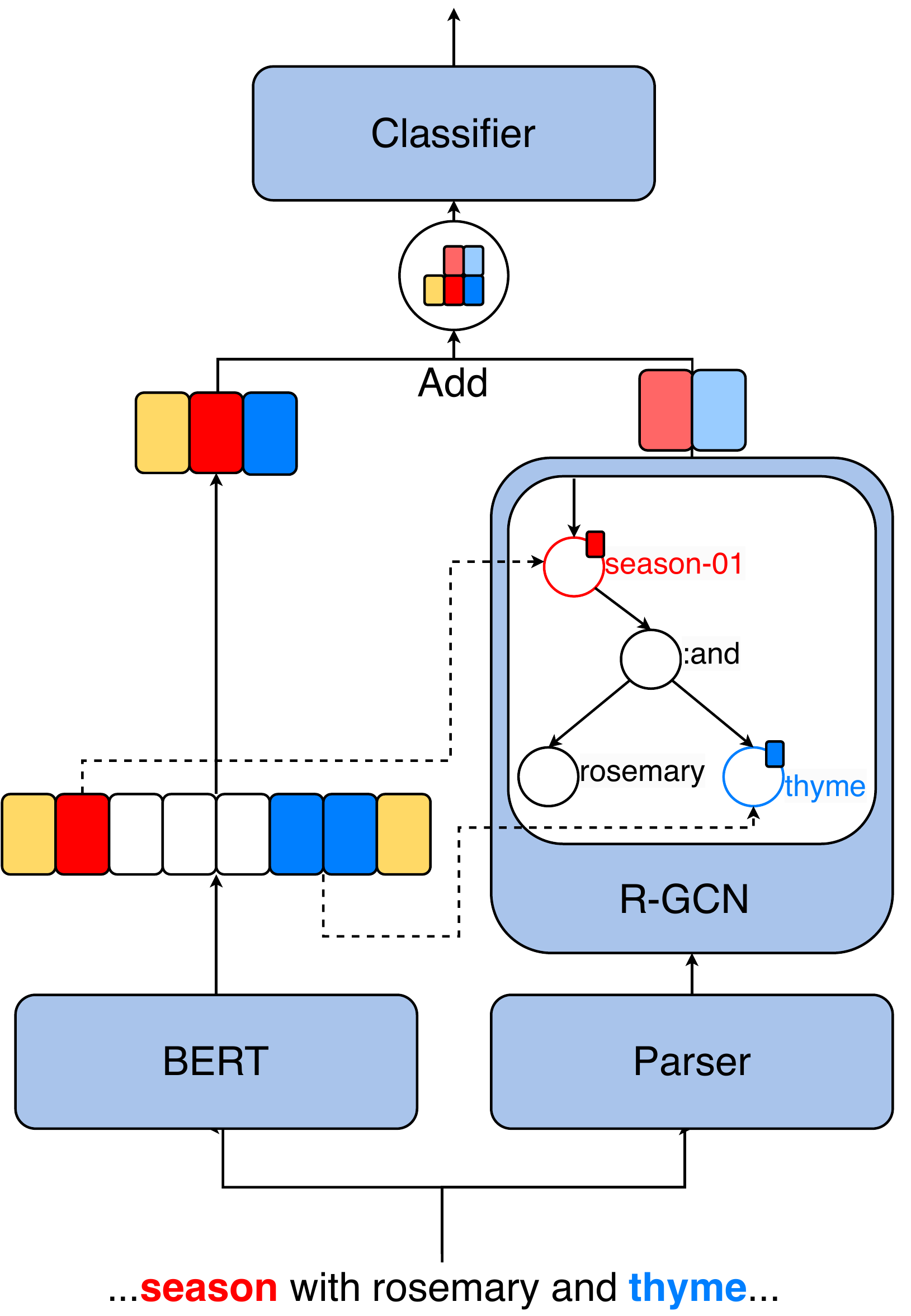}
    \caption{Model architecture. Yellow tokens denote BERT special tokens. Dotted lines indicate using BERT embeddings to seed the graph for the R-GCN.}
    \label{fig:arch}
\end{figure}

We design our methodology to test whether the inclusion of AMRs and dependency parses can improve the few-shot RE performance across datasets, by incorporating features from linguistic representations. We show an overview of our architecture in Figure \ref{fig:arch}, and go into further detail in Section \ref{sec:arch}. Our three datasets have their goal of generating a complete graph representation of a specified procedure. This graph is constructed by first finding salient entities in the procedural text, and then correctly identifying the appropriate relations between them. While this joint task is both challenging and useful, we restrict ourselves to the RE task for two reasons. Firstly, entity recognition results, as measured by baselines proposed in each of the dataset papers, vary widely, and entity recognition accuracy imposes an upper bound on end-to-end relation classification. Secondly, RE presents a common way to frame the tasks in each of these datasets. 

\subsection{Dataset Preprocessing} 
\label{sec:preprocessing}
In order to simplify our dataset tasks into relation extraction, we begin by identifying tuples of (entity1, relation, entity2), where each entity refers to a span of text in the original document, and relation refers to the flow graph edge label from the dataset. We format each triple into an instance that contains the triple and its context. We consider the context to be the shortest set of contiguous sentences that span both entity text spans. To segment sentences, we use the \texttt{en-core-sci-md} model with default settings provided in SciSpacy \citep{neumann-etal-2019-scispacy}, to account for the scientific text in the MSCorpus dataset. So that our models do not learn shallow heuristics to predict relations based on entity type, as observed in \citet{rosenman-etal-2020-exposing}, we exclude the entity types from the original datasets.

\subsection{Parsing}

We then annotate each context entity with two linguistic representations: AMR \citep{amr} and dependency parses. We choose AMR primarily for the quality of parsers available relative to other semantic formalisms: AMR parsing is a relatively popular task, and state-of-the-art parsers are often exposed to scientific text in their training. However, despite the quality of parses, AMR as a formalism presents several challenges to its use in downstream applications. Foremost among these is the problem of \textit{token alignment}: nodes and edges in AMR graphs do not have links back to the words in the text that they are generated from. As a contrast, we choose to use dependency parses as our syntactic framework, which are straightforward in their correspondence to the original text: each node corresponds to a word. 

For the dependency parses, we annotate each context span using the Stanza dependency parser \citep{qi2020stanza}, which produces a dependency graph per sentence. We then create a "top" node for the graph to link the individual trees for relations that span sentences. 

For the AMR parses, we use the SPRING model \citep{spring-amr} as implemented in AMRLib \footnote{\url{https://github.com/bjascob/amrlib}} We additionally verified that the model did not perform significantly differently than the original implementation. In contrast to the dependency parser, we found SPRING to occasionally be brittle. Because of its sequence-to-sequence architecture which cannot enforce that the produced output is a valid parse, the model sometimes failed to produce a parse altogether. These errors were non-transient, and did not display a pattern we could discern. In the interest of evaluating the impact of off-the-shelf tools as they were, we chose to include instances without AMR parses in our datasets. Because of the brittleness of the SPRING model, we parsed sentences in the datasets individually. We then compose the graph representations of  each context instance by joining the graphs of its constituent sentences. We follow the same procedure as with dependency parsing, joining all of the sentence-level AMR graphs with a top node. 

\subsection{AMR Alignment}

Because AMR nodes are not required to point back to the tokens that generated them, extracting token-level features to incorporate into our RE model relied on the task of AMR aligment. AMR alignment is usually treated as a \textit{post-hoc} task that relies on rule-based algorithms. We experimented with algorithms based on the common JAMR \citep{flanigan-etal-2014-discriminative} and ISI \citep{pourdamghani-etal-2014-aligning} aligners. These were implemented in AMRLib as the \texttt{RBW} and \texttt{FAA} aligners, respectively. Both aligners perform poorly, especially on the scientific text in the MSCorpus dataset. Because alignments are necessary to producing token-level features from an AMR representation, we developed heuristics as a second pass of alignment after applying the FAA aligner to the original text/AMR pair. Our heuristics, developed on the training split of each of our datasets, iteratively seek out unaligned AMR triples, normalize the node labels, and compare them with words in the original sentence after lemmatization. The words are taken from SPRING's tokenization of the original sentence, and the lemmatization uses NLTK's \citep{nltk} \texttt{WordNetLemmatizer} with the default parameters. We also normalize node labels to remove artifacts like Propbank sense indicators.

To measure the success of our alignment algorithm, we use a statistic that describes how many AMR triples in the graph that should be aligned (according to a combination of the AMR standard\footnote{https://amr.isi.edu/doc/amr-alignment-guidelines.html} and dataset-specific heuristics), are aligned to a token in the text. We also compute statistics based on how many triples contain at least one entity unaligned with the graph. With only the FAA aligner, over 59\% of triples contain at least one entity without a corresponding aligned word across our three datasets. After realignment, we achieve a significantly higher rate of alignment, with just under 27\% of triples having at least one entity unaligned to nodes in the graph. 

\subsection{Model Architectures}
\label{sec:arch}
\textbf{Baseline Model:} We consider a common baseline architecture for relation extraction, based on BERT \citep{devlin-etal-2019-bert}. We begin by embedding the context for each relation. We then extract the embeddings for all tokens that constitute each entity, and max-pool them into embeddings $e_1$ and $e_2$ . We concatenate $e_1$, $e_2$, and the embedding for the \texttt{[CLS]} token, which we consider a stand-in for the context, into one vector. We then pass that vector through a two-layer MLP with a $\tanh$ activation between layers, before finally applying a softmax for the classification. 

\textbf{Graph-aware models:} To compute graph-based features, we first initialize the linguistic graph's nodes with feature vectors of the same size as the baseline BERT model's embeddings. For every aligned token, we initialize that feature vector with the max-pool of the embeddings of each of its aligned tokens, leaving the embeddings zeroed out for unaligned nodes. We then pass the graph through a relational graph convolution network (R-GCN, \citet{rgcn}). We choose the R-GCN for its ability to model heterogeneous relations in graphs. After computing node embeddings, we employ a residual connection similar to the \texttt{Hier} setting shown in figure 3a in \citet{bai-etal-2021-semantic}, where the mean pool of node embeddings corresponding to $e_1$ and $e_2$ is added back to the BERT-based embeddings of the aligned entity tokens computed earlier. These updated embeddings are then passed to the same MLP relation classifier as in the baseline. We choose this type of residual connection for the bottleneck in representational capacity that it imposes on our models. Additionally, we measure the distribution of path lengths between entities in both frameworks in the train split of our datasets, and find that the mean path of each dataset lies between 3 and 4. We thus use an R-GCN of depth 4 for all experiments in order to capture most paths. Because of the residual connection architecture, we are restricted to using the baseline BERT model's word embedding size as the node embedding size as well. Combined with the GNN depth of 4, our model adds significantly more parameters — 203M parameters vs the plaintext model's 111M. However, we hypothesize that being forced to operate in the same embedding space as the baseline will discourage models from memorizing the original dataset and overfitting, especially in the few-shot setting. 

We depict our architecture in figure \ref{fig:arch}. The baseline architecture omits the right-hand fork, using only BERT embeddings.

\section{Datasets}

We consider three datasets across two different domains for this transfer: cooking and materials science procedures. Our cooking datasets are the RISeC \citep{risec} and English Recipe Flow Graph (EFGC) \citep{yamakata-etal-2020-english} corpora, and we introduce a much wider domain gap with the Materials Science Procedural Text Corpus (MSCorpus) from \citet{mscorpus}. We do not standardize labels across datasets; we retain the original labels from each dataset, though we combine some relations in MSCorpus to make it more comparable to the other datasets (see below for details). Summary statistics for each dataset (including for the definition of "relation" described in section \ref{sec:preprocessing}) are shown in table \ref{tab:datastats}, and we describe salient features for each of our datasets below. Notably, all three of our datasets exhibit a high degree of concentration in their label distributions, with infrequent classes being found sometimes as much as 200$\times$ less than the most frequent classes.

\begin{table}[h]
\resizebox{\columnwidth}{!}{
    \begin{tabular}{lllll}
        \textbf{Dataset} & \textbf{Documents} & \textbf{\# Relations} & \textbf{Labels} & \textbf{Label Distribution} \\ \hline 
        RISeC            & 260                & 7,591 &   11        & \includegraphics[width=3cm]{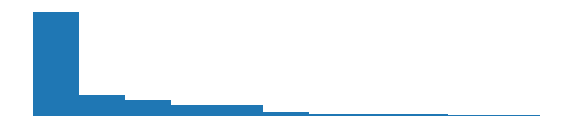}                          \\ 
        EFGC             & 300                & 15,681  &  13      & \includegraphics[width=3cm]{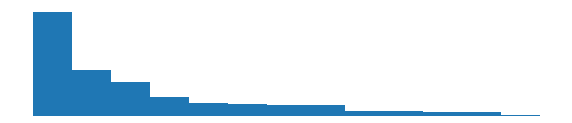}                             \\ 
        MSCorpus         & 230                & 18,399  &  11     &    \includegraphics[width=3cm]{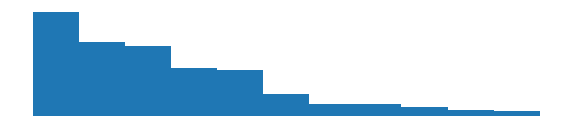}                         \\ \hline
    \end{tabular}
}
\caption{Dataset Statistics. The label distribution column visualizes sorted frequencies of labels in each dataset.}
\label{tab:datastats}
\end{table}

The \textbf{RISeC} dataset \citep{risec} is the most explicitly aligned with existing semantic frameworks: the authors build upon Propbank \citep{propbank}, which is also the framework that underlies AMR. However, because the relations in the dataset do not correspond strictly to verbal frames, relations use Propbank roles, rather than numbered arguments. Additionally, while these relations are \textit{inspired} by Propbank, the authors' definitions of the labels do not always correspond to Propbank's, rendering this correspondence somewhat loose. 

The \textbf{EFGC} dataset takes a more domain-specific approach, and defines a labeling schema specialized for cooking, including coreference relations segmented by whether the coreferent entities are tools, foods, or actions. Many of the descriptors of actions that are given explicit labels in RISeC such as temporal relations and descriptions of manner, are collapsed into a single class in this dataset, with the authors choosing to focus on physical components, their amounts, and operational relationships. 

The \textbf{MSCorpus} dataset splits its relations into three categories: relations between operations and entities, relations between entities, and one relation indicating the flow of operations. MSCorpus defines a rich set of relations between entites, which is atypical for the other datasets. We thus combine some of these labels to bring MSCorpus into alignment with the other annotation schemas. 

\section{Experiments}

\subsection{In-Domain Experiments}

We train both the baseline and graph-aware models on each dataset, using the train/dev/test splits where provided. If no dev split was provided, we randomly split the training dataset 80/20 into new train and dev splits.
We use \texttt{bert-base-uncased} as available on the Huggingface Hub \footnote{\url{https://huggingface.co/bert-base-uncased}} as our base BERT model, for both the baseline and graph-aware variants. For our graph-aware variants, we use R-GCN as our graph network. We train each model with the Adam optimizer \citep{Kingma2014AdamAM} to minimize the cross-entropy loss between predicted and true labels. We use a learning rate of $2\times10^{-5}$ and a batch size of 16. Each model is trained on 3 random seeds for 30 epochs, using early stopping criterion based on the macro-averaged F1 score on the dev split with a patience of 5 epochs. We keep the model that performs best on the dev split, and calculate its corresponding macro F1 score on the test set. We refer to the graph aware models that add dependencies and AMRs as +Dep and +AMR, respectively.

\subsection{Few-shot Experiments}
We formulate few-shot transfer learning as an $N$-way $K$-shot problem, where a model is trained on $K$ instances of each of the $N$ classes in the target domain. We experiment with $K \in \{1, 5, 10, 20, 50, 100\}$. Because of the label imbalance in our datasets, where $K$ is greater than the number of labeled examples for a given class, we sample all of the labeled instances without replacement. This can result in fewer than K examples for a given class.

For the transfer process, we begin with the models trained in the in-domain experiments, and replace the MLP classification head with a  freshly initialized head with a suitable number of outputs for the target domain's number of classes. We reuse the BERT and R-GCN components of the in-domain model, and allow their weights to be updated in the transfer finetuning. 

We continue to train each model using the same settings as in-domain training using a batch size of 4, sampling each dataset three times with different seeds. 

In addition, to control for the effects of the source and target dataset interactions and our sampling strategies, we train few-shot models in each domain from scratch, for each of the settings $K$ described above, using the same settings. 

All of our experiments were run on NVIDIA A4500 GPUs, and we used roughly 33 days of GPU time for all of the experiments in this project, including hyperparameter tuning. 

\section{Results and Discussion}





\begin{figure*}[ht]
    \centering
    \begin{subfigure}[b]{0.45\textwidth}
        \includegraphics[width=\textwidth]{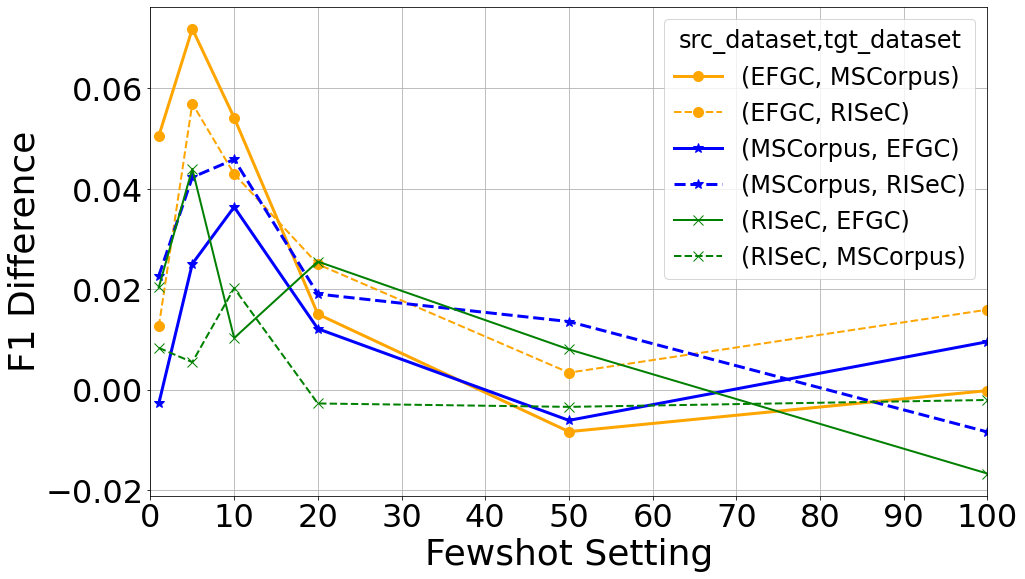}
        \caption{+AMR vs baseline}
    \end{subfigure}
    \begin{subfigure}[b]{0.45\textwidth}
        \includegraphics[width=\textwidth]{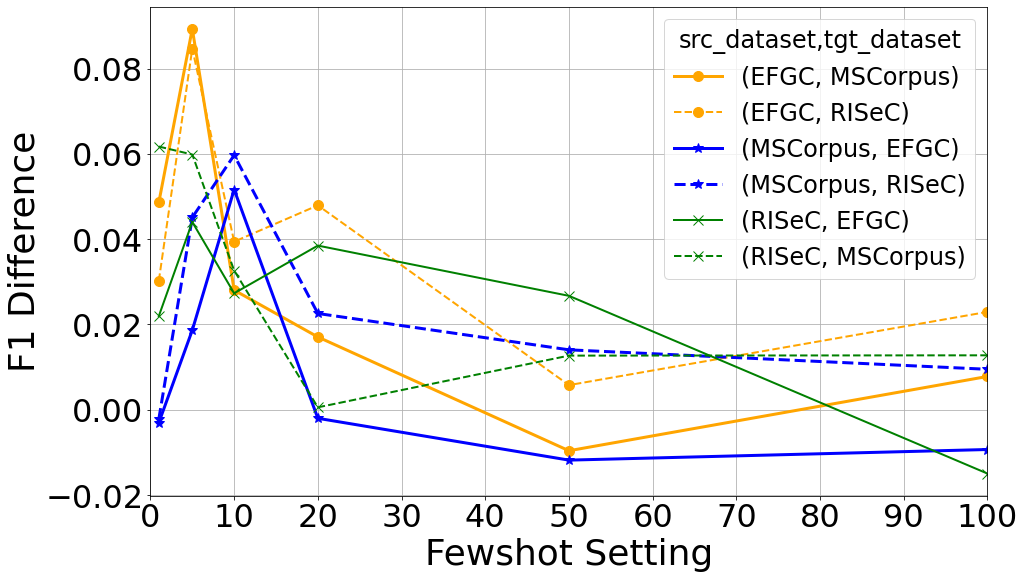}
        \caption{+Dep vs baseline}
    \end{subfigure}
    \caption{Differences in F1 over baseline from incorporating linguistic graphs in models.}
    \label{fig:diff_curve}
\end{figure*}
We expect that more powerful linguistic representations than plain text will aid in few shot transfer between domains.  In order for few shot transfer to be successful, the target data points used for transfer need to increase the relevant shared representation between the source and target datasets.  Because of this, we expect that any effect of representation on test set performance will depend upon how much shared representation there was between the two domains to begin with and how much the few added examples closes the gap. A more efficient representation may lose its advantage once there are enough target domain examples to obviate the need for efficiency. In this section, we aim to answer a number of questions.

\begin{table}[h]
\centering
\begin{tabular}{lll}
\toprule
\textbf{Dataset} & \textbf{Case} &      \textbf{Mean (std)}                                  \\
\midrule
EFGC & +AMR &  83.9 \small{ \color{gray}(0.3)} \\
      & +Dep &  84.6 \small{ \color{gray}(1.3)} \\
      & Baseline &  85.0 \small{ \color{gray}(0.8)} \\ \hline
MSCorpus & +AMR &  87.8 \small{ \color{gray}(1.0)} \\
      & +Dep &  88.4 \small{ \color{gray}(0.5)} \\
      & Baseline &  87.5 \small{ \color{gray}(0.5)} \\ \hline
RISeC & +AMR &  82.8 \small{ \color{gray}(1.6)} \\
      & +Dep &  81.7 \small{ \color{gray}(2.1)} \\
      & Baseline &  82.7 \small{ \color{gray}(1.3)} \\ \hline
\bottomrule
\end{tabular}

\caption{Results from in-domain experiments. Each value represents the mean of runs with three random seeds, with standard deviation in parentheses.}
\label{tab:indomain}
\end{table}
\paragraph{Do linguistic representation aid in either in-domain or cross-domain transfer?}
We present our in-domain results on the complete datasets in table \ref{tab:indomain}. Overall, we do not see significant differences between the baseline and +Dep and +AMR cases, even though they appear to overperform the baseline case on RISeC and MSCorpus. Notably, however, these models do not overfit more than the baseline: performance on the unseen test set remains similar. 

We do, however, see differences in performance between the baseline and graph-aware cases in the few-shot transfer setting. In Figure \ref{fig:diff_curve}, we visualize the difference between the macro-averaged F1 performance in each of our graph-aware cases and the baseline against the few-shot setting. We see that while in the 1-shot case, our results are highly variable, the 5-, 10-, and 20- shot cases yield noticeable improvement, peaking in the 5- and 10-shot settings. In our best-performing results, we see a 6-point absolute gain in F1 score.

\begin{table*}[ht]
\centering
    \resizebox{0.9\textwidth}{!}{
\begin{tabular}{lllllllll}
\toprule
     &          &      \multicolumn{6}{c}{\textbf{Fewshot Setting}} \\
\textbf{Target} & \textbf{Source} & \textbf{Case} &                                         1   &                                        5   &                                        10  &                                        20  &                                        50  &                                        100 \\
\midrule
RISeC & From Scratch & Baseline &           18.6 \small{ \color{gray} (2.9)} &           36.5 \small{ \color{gray} (3.2)} &           48.3 \small{ \color{gray} (3.1)} &           60.2 \small{ \color{gray} (2.3)} &  \textbf{71.1 \small{ \color{gray} (1.1)}} &           76.9 \small{ \color{gray} (0.2)} \\
     &          & +Dep &           19.3 \small{ \color{gray} (4.5)} &  \textbf{40.0 \small{ \color{gray} (2.8)}} &           51.5 \small{ \color{gray} (3.2)} &  \textbf{62.7 \small{ \color{gray} (4.5)}} &           71.1 \small{ \color{gray} (0.9)} &  \textbf{79.5 \small{ \color{gray} (1.5)}} \\
     &          & +AMR &  \textbf{19.8 \small{ \color{gray} (7.1)}} &           39.3 \small{ \color{gray} (5.2)} &  \textbf{52.1 \small{ \color{gray} (2.6)}} &           60.9 \small{ \color{gray} (4.0)} &           70.6 \small{ \color{gray} (0.7)} &           78.4 \small{ \color{gray} (1.0)} \\ \hline
     & MSCorpus & Baseline &           19.7 \small{ \color{gray} (5.5)} &           35.1 \small{ \color{gray} (5.4)} &           45.6 \small{ \color{gray} (0.8)} &           57.7 \small{ \color{gray} (0.9)} &           67.8 \small{ \color{gray} (1.3)} &           76.2 \small{ \color{gray} (1.6)} \\
     &          & +Dep &           19.4 \small{ \color{gray} (2.1)} &  \textbf{39.7 \small{ \color{gray} (5.2)}} &  \textbf{51.6 \small{ \color{gray} (1.1)}} &  \textbf{60.0 \small{ \color{gray} (4.8)}} &  \textbf{69.2 \small{ \color{gray} (1.9)}} &  \textbf{77.2 \small{ \color{gray} (3.4)}} \\
     &          & +AMR &  \textbf{21.9 \small{ \color{gray} (2.6)}} &           39.4 \small{ \color{gray} (4.2)} &           50.2 \small{ \color{gray} (0.9)} &           59.6 \small{ \color{gray} (0.9)} &           69.2 \small{ \color{gray} (2.2)} &           75.4 \small{ \color{gray} (1.3)} \\ \hline
     & EFGC & Baseline &           25.8 \small{ \color{gray} (5.0)} &           42.0 \small{ \color{gray} (4.0)} &           53.7 \small{ \color{gray} (0.6)} &           61.8 \small{ \color{gray} (3.3)} &           71.1 \small{ \color{gray} (1.5)} &           75.2 \small{ \color{gray} (0.9)} \\
     &          & +Dep &  \textbf{28.8 \small{ \color{gray} (7.7)}} &  \textbf{50.5 \small{ \color{gray} (3.9)}} &           57.6 \small{ \color{gray} (2.4)} &  \textbf{66.6 \small{ \color{gray} (0.9)}} &  \textbf{71.7 \small{ \color{gray} (1.2)}} &  \textbf{77.5 \small{ \color{gray} (0.8)}} \\
     &          & +AMR &           27.0 \small{ \color{gray} (7.6)} &           47.7 \small{ \color{gray} (8.9)} &  \textbf{58.0 \small{ \color{gray} (2.6)}} &           64.3 \small{ \color{gray} (1.2)} &           71.5 \small{ \color{gray} (0.4)} &           76.8 \small{ \color{gray} (2.1)} \\ \hline
MSCorpus & From Scratch & Baseline &           25.0 \small{ \color{gray} (4.9)} &           46.9 \small{ \color{gray} (2.7)} &           63.4 \small{ \color{gray} (1.0)} &  \textbf{74.0 \small{ \color{gray} (1.1)}} &           82.7 \small{ \color{gray} (1.2)} &           82.6 \small{ \color{gray} (1.9)} \\
     &          & +Dep &  \textbf{30.6 \small{ \color{gray} (2.8)}} &  \textbf{49.5 \small{ \color{gray} (1.0)}} &  \textbf{66.0 \small{ \color{gray} (3.2)}} &           72.7 \small{ \color{gray} (2.3)} &  \textbf{82.7 \small{ \color{gray} (0.9)}} &  \textbf{84.8 \small{ \color{gray} (0.3)}} \\
     &          & +AMR &           26.7 \small{ \color{gray} (4.3)} &           45.3 \small{ \color{gray} (0.9)} &           62.4 \small{ \color{gray} (3.1)} &           72.6 \small{ \color{gray} (2.0)} &           82.2 \small{ \color{gray} (1.2)} &           84.3 \small{ \color{gray} (1.0)} \\ \hline
     & RISeC & Baseline &           24.4 \small{ \color{gray} (2.2)} &           43.4 \small{ \color{gray} (2.5)} &           56.5 \small{ \color{gray} (3.3)} &           69.8 \small{ \color{gray} (1.3)} &           81.4 \small{ \color{gray} (0.9)} &           83.7 \small{ \color{gray} (0.6)} \\
     &          & +Dep &  \textbf{30.6 \small{ \color{gray} (0.5)}} &  \textbf{49.4 \small{ \color{gray} (3.5)}} &  \textbf{59.8 \small{ \color{gray} (3.9)}} &  \textbf{69.9 \small{ \color{gray} (4.2)}} &  \textbf{82.6 \small{ \color{gray} (1.0)}} &  \textbf{85.0 \small{ \color{gray} (1.4)}} \\
     &          & +AMR &           25.3 \small{ \color{gray} (3.1)} &           43.9 \small{ \color{gray} (3.4)} &           58.5 \small{ \color{gray} (4.9)} &           69.5 \small{ \color{gray} (2.2)} &           81.0 \small{ \color{gray} (1.0)} &           83.5 \small{ \color{gray} (1.5)} \\ \hline
     & EFGC & Baseline &           26.9 \small{ \color{gray} (4.6)} &           46.6 \small{ \color{gray} (2.1)} &           63.8 \small{ \color{gray} (3.0)} &           72.5 \small{ \color{gray} (0.9)} &  \textbf{81.5 \small{ \color{gray} (0.9)}} &           83.6 \small{ \color{gray} (1.8)} \\
     &          & +Dep &           31.7 \small{ \color{gray} (4.0)} &  \textbf{55.5 \small{ \color{gray} (5.6)}} &           66.6 \small{ \color{gray} (4.6)} &  \textbf{74.2 \small{ \color{gray} (2.5)}} &           80.5 \small{ \color{gray} (3.0)} &  \textbf{84.4 \small{ \color{gray} (1.1)}} \\
     &          & +AMR &  \textbf{31.9 \small{ \color{gray} (3.8)}} &           53.8 \small{ \color{gray} (6.1)} &  \textbf{69.3 \small{ \color{gray} (0.8)}} &           74.0 \small{ \color{gray} (3.3)} &           80.7 \small{ \color{gray} (1.2)} &           83.6 \small{ \color{gray} (2.1)} \\ \hline
EFGC & From Scratch & Baseline &           16.2 \small{ \color{gray} (1.5)} &           29.3 \small{ \color{gray} (2.3)} &           38.9 \small{ \color{gray} (1.8)} &           47.6 \small{ \color{gray} (1.2)} &           61.0 \small{ \color{gray} (0.9)} &           63.8 \small{ \color{gray} (3.0)} \\
     &          & +Dep &  \textbf{17.2 \small{ \color{gray} (4.1)}} &           30.3 \small{ \color{gray} (3.8)} &  \textbf{40.7 \small{ \color{gray} (2.5)}} &           48.6 \small{ \color{gray} (1.1)} &           60.2 \small{ \color{gray} (1.8)} &  \textbf{66.7 \small{ \color{gray} (2.4)}} \\
     &          & +AMR &           14.2 \small{ \color{gray} (2.3)} &  \textbf{30.8 \small{ \color{gray} (2.2)}} &           39.9 \small{ \color{gray} (3.3)} &  \textbf{48.7 \small{ \color{gray} (1.1)}} &  \textbf{61.1 \small{ \color{gray} (2.1)}} &           64.1 \small{ \color{gray} (3.2)} \\ \hline
     & RISeC & Baseline &           16.0 \small{ \color{gray} (1.7)} &           30.4 \small{ \color{gray} (3.0)} &           35.7 \small{ \color{gray} (0.4)} &           44.7 \small{ \color{gray} (1.5)} &           56.9 \small{ \color{gray} (1.2)} &  \textbf{65.8 \small{ \color{gray} (1.6)}} \\
     &          & +Dep &  \textbf{18.2 \small{ \color{gray} (4.5)}} &  \textbf{34.8 \small{ \color{gray} (3.0)}} &  \textbf{38.4 \small{ \color{gray} (3.2)}} &  \textbf{48.6 \small{ \color{gray} (1.3)}} &  \textbf{59.5 \small{ \color{gray} (1.6)}} &           64.3 \small{ \color{gray} (2.9)} \\
     &          & +AMR &           18.1 \small{ \color{gray} (1.5)} &           34.8 \small{ \color{gray} (1.4)} &           36.7 \small{ \color{gray} (1.5)} &           47.3 \small{ \color{gray} (2.4)} &           57.7 \small{ \color{gray} (2.7)} &           64.1 \small{ \color{gray} (2.9)} \\ \hline
     & MSCorpus & Baseline &  \textbf{17.4 \small{ \color{gray} (4.4)}} &           29.5 \small{ \color{gray} (2.9)} &           39.7 \small{ \color{gray} (2.8)} &           49.2 \small{ \color{gray} (0.5)} &  \textbf{61.2 \small{ \color{gray} (1.0)}} &           64.4 \small{ \color{gray} (1.0)} \\
     &          & +Dep &           17.0 \small{ \color{gray} (3.8)} &           31.4 \small{ \color{gray} (2.2)} &  \textbf{44.9 \small{ \color{gray} (1.9)}} &           49.0 \small{ \color{gray} (1.2)} &           60.0 \small{ \color{gray} (0.6)} &           63.5 \small{ \color{gray} (3.4)} \\
     &          & +AMR &           17.1 \small{ \color{gray} (2.4)} &  \textbf{32.0 \small{ \color{gray} (0.2)}} &           43.4 \small{ \color{gray} (2.4)} &  \textbf{50.4 \small{ \color{gray} (2.6)}} &           60.6 \small{ \color{gray} (0.6)} &  \textbf{65.4 \small{ \color{gray} (3.7)}} \\
\bottomrule
\end{tabular}
}
\caption{Few-shot learning results. "From Scratch" in the source column represents the case where we train a few-shot model from scratch, without transfer. Each cell represents the mean macro-F1 across three random seeds, with the standard deviation of those runs in parentheses. We group our results by the target dataset first to allow easier comparison of the impact of source datasets. Bold results represent the best case for a source-target pair.}
\label{tab:all_fewshot_results}

\end{table*}

We find that both dependency parse and AMR representations show a statistically significant positive effect on performance. In particular, we test the significance of the effect with an ANOVA model with multiple independent variables: namely, source and target dataset (EFGC, RISeC, MSCorpus), representation case (Baseline, +Dep, +AMR), few-shot setting (1, 5, 10, 20, 50, 100), and transfer setting (in-domain vs out-of-domain). The dependent variable is test set F1. The data table for the analysis includes 3 runs for each combination of variables each with a separate random seed. We train our models under a full-factorial experimental design, i.e. we ran trials for all combinations of variables. This design allows us to test the reliability of the effect of our variables under a variety of conditions while making the necessary statistical adjustments to avoid spurious significant effects that may occur when multiple statistical comparisons are made. We use this design rather than pairwise significance tests so that we can measure the effect of introducing linguistic formalisms as a whole, rather than arguing the statistical significance of individual, pairwise comparisons.

We expect that the similarity between source and target datasets, the variation in the target dataset, and the few-shot setting could all either dampen or magnify any effect of representation on the performance. We therefore include pairwise interaction terms in the ANOVA model for case by source dataset, case by target dataset, case by transfer setting, and case by few-shot setting. The examples added for the few shot setting in the transfer case are sampled from the training split of the target dataset. Thus, while we expect for the cross-domain case the few shot setting has an effect, we do not expect an effect in the in-domain case, since the target domain examples added to the training data simply replicate examples that were already part of the dataset. To account for this, we include a final interaction term between few shot setting and transfer setting in the ANOVA model.

The ANOVA model explains 98\% of the variation in F1 scores.  The results align well with our intuitions. First, as expected we find a significant effect of transfer setting such that in-domain performance on the entire dataset is better than transfer performance in a few-shot setting: F(1, 679) = 10356.25, p < .0001. In these cases, the original dataset for in-domain training is between 5 and 15 times the size of the target training dataset.  We also find a significant effect of the few-shot setting, such that larger numbers of target domain examples are associated with higher performance, F(5, 679) = 716.79, p < .0001.  A post-hoc student-t analysis reveals that all pairwise comparisons are significant.  Notably, there is a significant interaction between transfer setting and few shot setting: F(5, 679) = 733.83, p < .0001, such that the effect of the few-shot setting is restricted to the transfer setting, as expected. 

Our hypothesis is primarily related to the importance of the representation of the data for efficiently enabling transfer between domains.  We find a significant effect of representation case: F(2, 679) = 5.26, p < .01.  \textbf{A student-t post-hoc analysis reveals that both +Dep and +AMR cases are better than plain text, but there is no significant difference between the two.}  There is also a significant interaction between representation case and transfer setting: F(2, 679) = 8.19, p < .0005.  In particular, the effect of case is only significant in the transfer setting.  There is also a significant interaction between few shot setting and case: F(2, 679) = 8.19, p < .0005.  A student-t post-hoc analysis reveals that the effect is only significant for the 5- and 10-shot settings.  Thus, \textbf{1 target example is too small to yield a significant effect whereas 20 or more is too many such that the representational advantage disappears}.  We also find a significant interaction between representation case and target dataset, but not with source dataset: F(4, 679) = 2.61, p < .05, such that the effect of representation is significant for RISeC and MSCorpus but not for EFGC.

We present all of our few-shot results in Table \ref{tab:all_fewshot_results}. Significance testing was performed on the difference in results between the baseline and lingusitic representation cases in the transfer setting. Additionally, we investigate the impact of  source domain on the utility of linguistic representations. We therefore compare results between models trained in a few-shot setting from scratch, seeing only one dataset, with the transfer model that we train on a source dataset first. We show both of these cases in table \ref{tab:all_fewshot_results}, with few-shot models trained from scratch denoted in the source dataset column as "From Scratch" results. 

\textbf{How important is the choice of the source domain on the transfer performance?} We see several interesting patterns in our 5- and 10-shot results when we take our few-shot models trained from scratch into account. We visualizes differences in performance between the from-scratch models and models trained with a different source domain in table \ref{tab:diff_from_scratch}. We find that the transfer between datasets for our text-only models is of limited utility, if not outright harmful. While we see one instance (the EFGC to RISeC transfer) in which introducing a transfer source dataset improves the baseline model's performance on the target dataset consistently, we see more commonly that adding a transfer source dataset makes only a small difference, or even hurts the performance of the baseline model. In the cases of transfer between MSCorpus and RISeC in either direction, for instance, the baseline model in the transfer setting consistently underperforms the model trained from scratch by up to 7 F1 points, and does not close that gap even in the 50- and 100-shot settings. However, incorporating linguistic formalisms proves to be far more robust to the choice of source domain: the linguistic representations, regardless of source domain are never worse than the baseline trained on that source domain, and still frequently outperform the baseline trained from scratch, even when the choice of source domain imposes a performance penalty. 

\begin{table}[ht]
\centering
    \resizebox{0.9\columnwidth}{!}{
    \begin{tabular}{lllll}
\toprule
    &          &      \multicolumn{3}{c}{\textbf{Fewshot Setting}} \\
\textbf{Target} & \textbf{Source} & \textbf{Case} & 5   &  10  \\
\midrule
RISeC & MSCorpus & Baseline &  -1.37 &  -2.66 \\
     &          & +Dep &  -0.29 &   0.04 \\
     &          & +AMR &   0.10 &  -1.92 \\ \hline
     & EFGC & Baseline &   5.50 &   5.42 \\
     &          & +Dep &  10.53 &   6.09 \\
     &          & +AMR &   8.44 &   5.87 \\  \hline
MSCorpus & RISeC & Baseline &  -3.55 &  -6.92 \\
     &          & +Dep &  -0.10 &  -6.24 \\
     &          & +AMR &  -1.40 &  -3.91 \\  \hline
     & EFGC & Baseline &  -0.33 &   0.41 \\
     &          & +Dep &   6.06 &   0.63 \\
     &          & +AMR &   8.45 &   6.82 \\  \hline
EFGC & RISeC & Baseline &   1.12 &  -3.23 \\
     &          & +Dep &   4.51 &  -2.34 \\
     &          & +AMR &   4.02 &  -3.23 \\  \hline
     & MSCorpus & Baseline &   0.29 &   0.85 \\
     &          & +Dep &   1.14 &   4.18 \\
     &          & +AMR &   1.29 &   3.46 \\
\bottomrule
\end{tabular}
}
\caption{Differences from baseline model trained from scratch in the 5- and 10-shot cases gained in using a different source domain. Linguistic representations are more robust to choice of source domain.}
\label{tab:diff_from_scratch}
\end{table}

Interestingly, an intuitive notion of "domain distance" fails to explain when transfer will be helpful. EFGC and RISeC both come from the cooking domain, but though RISeC and MSCorpus negatively influence each other in transfer, MSCorpus and EFGC in the baseline case have very little difference from the transfer case. Transfer between the abstract categories of "cooking" dataset and "materials science" dataset is highly variable. 

Notably, we observe that the benefits we derive from transfer seem asymmetrical: even datasets that transfer well in one direction might not in the other direction. We see markedly better results transferring from EFGC to RISeC, for instance, than we see in the reverse direction, and we see a similar result (though less consistent) for transfer from EFGC to MSCorpus as compared to the reverse. 

\textbf{What is the impact of linguistic structure on the  performance of few-shot RE in-domain?} When factoring in the effect of our graph-aware models, we see that they help models generalize, both in the few-shot in-domain setting, as well as the transfer setting. \textbf{Where transfer itself causes the performance of the baseline model to degrade, however, we see that the addition of linguistic representations sometimes makes up for that gap almost entirely}. In the case of the 10-shot MSCorpus to RISeC transfer, we see that the baseline transfer model performs an average 2.7 points worse than the baseline from-scratch model (48.5 vs. 45.3), but that the dependency models perform very similarly (51.5 vs. 51.6). In cases where the transfer pairs are well-matched, however, we see that while the baseline results remain similar, the benefit that the models derive from the linguistic representations is much more pronounced in the transfer setting. In the 10-shot transfer in both directions between EFGC and MSCorpus, as well as the EFCG to RISeC case, transfer models that incorporate dependencies and AMRs overperform their in-domain counterparts by between 3 and 7 points.

\section{Conclusion and Future Work}

We experiment with using linguistic formalisms as additional context for learning robust representations that facilitate few-shot transfer among domains for the task of relation extraction. Our experiments show that the inclusion of linguistic formalisms significantly boosts models' ability to transfer to new datasets. They additionally show that that benefit is robust to whether transfer learning helps in the baseline case. This suggests that using linguistic formalisms as a scaffold for learning in data scarce, specialized domains could be a powerful technique. 

Future work could focus on several directions. With regards to the use of semantic frameworks, more work is needed to understand how best to incorporate highly abstract formalisms such as AMR. For example, how can we better use the node features in AMR, rather than just the structure? These questions also apply to more abstract syntactic frameworks like constituency parsing. With regards to our transfer learning process, we aim to understand what features of a pair of datasets make them suited to transferring by studying a wider array of datasets in diverse domains, as well as to study the impact of domain adapting our syntactic and semantic parsers to our target domains.

\section*{Limitations}


Because we focus on off-the-shelf tools in this work, we are necessarily constrained by the availability of such tools in different languages and contexts. While dependency annotation tools are widely available for many languages through projects like the Universal Dependency project, semantic frameworks, let alone effective, accurate parsers for them, are harder to find. In addition, we are constrained by the current state of the art for AMR parsing and, more challengingly, alignment. AMR parsing continues to improve, but alignment has only recently attracted interest again as a problem, such as in \citep{x-attn-alignment}. 

Additionally, this work, in evaluating six few-shot settings across six pairs of datasets and a number of seeds suffers from a combinatorial problem in terms of the necessary compute infrastructure. As discussed in the paper, our work consumed roughly a month of GPU time. Combined with the size of the models, this limits the accessibility of this vein of research. More effort understanding how to narrow down the choice of datasets before studying transfer would go a long way towards alleviating this issue.

\section*{Acknowledgments}

This research was sponsored by the Defense Advanced Research Projects Agency and the Army Research Office and was accomplished under Grant Number W911NF-20-1-0006. The views and conclusions contained in this document are those of the authors and should not be interpreted as representing the official policies, either expressed or implied, of the Defense Advanced Research Projects Agency, the Army Research Office, or the U.S. Government. The U.S. Government is authorized to reproduce and distribute reprints for Government purposes notwithstanding any copyright notation herein.


\bibliography{anthology,custom}
\bibliographystyle{acl_natbib}

\end{document}